\documentclass[preprint,article,accept,moreauthors,pdftex]{Definitions/mdpi}

\usepackage{amssymb}
\usepackage{url}

\usepackage{hyperref}

\usepackage[utf8]{inputenc} 
\usepackage{booktabs}       
\usepackage{amsfonts}       
\usepackage{nicefrac}       
\usepackage{microtype}      
\usepackage{epsfig}
\usepackage{graphicx}
\usepackage{subfigure} 

\nolinenumbers

\firstpage{1} 
\pubvolume{xx}
\issuenum{1}
\articlenumber{5}
\pubyear{2020}
\copyrightyear{2020}
\history{Received: date; Accepted: date; Published: date}

\Title{Relation3DMOT: Exploiting Deep Affinity for 3D Multi-Object Tracking from View Aggregation}

\Author{Can Chen $^{1}$, Luca Zanotti Fragonara $^{1}$* and Antonios Tsourdos $^{1}$}

\AuthorNames{Can Chen, Luca Zanotti Fragonara and Antonios Tsourdos}

\address{%
$^{1}$ \quad School of Aerospace, Transport and Manufacturing, Cranfield University, UK, MK43 0AL; can.chen@cranfield.ac.uk; (C.C.); a.tsourdos@cranfield.ac.uk (A.T.)}

\corres{Correspondence: l.zanottifragonara@cranfield.ac.uk}

\abstract{Autonomous systems need to localize and track surrounding objects in 3D space for safe motion planning. As a result, 3D multi-object tracking (MOT) plays a vital role in autonomous navigation. Most MOT methods use a tracking-by-detection pipeline, which includes object detection and data association processing. However, many approaches detect objects in 2D RGB sequences for tracking, which is lack of reliability when localizing objects in 3D space. Furthermore, it is still challenging to learn discriminative features for temporally-consistent detection in different frames, and the affinity matrix is normally learned from independent object features without considering the feature interaction between detected objects in the different frames. To settle these problems, We firstly employ a joint feature extractor to fuse the 2D and 3D appearance features captured from both 2D RGB images and 3D point clouds respectively, and then propose a novel convolutional operation, named RelationConv, to better exploit the correlation between each pair of objects in the adjacent frames, and learn a deep affinity matrix for further data association. We finally provide extensive evaluation to reveal that our proposed model achieves state-of-the-art performance on KITTI tracking benchmark.}

\keyword{3D Multi-object Tracking; Sensor Fusion; Deep Affinity; Relation Learning; Neural Network}

\begin{document}


\section{Introduction}
Multi-object tracking in 3D world space (3D MOT) plays an indispensable role in environmental perception for autonomous systems \cite{luiten2020track, zhang2019robust, hu2019joint, weng20203d}. The 3D MOT task is that it firstly localizes all the surrounding 3D objects in a sequence, and then assigns them the consistent identity numbers (IDs). As a result, the same objects in the sequence are given the same IDs, which is then used to predict the trajectories for the surrounding objects and make safe path planning for autonomous navigation. In recent years, 2D multi-object tracking has made great progress in the computer vision domain \cite{bergmann2019tracking, zhou2020tracking, sun2019deep, wang2019towards, milan2016online}. However, the camera sensor is unlikely to provide depth information and is quite sensitive to the lighting conditions (e.g. overexposure) for the autonomous systems in the real 3D world.

In order to improve the reliability and safety, recent tracking-by-detection approaches \cite{zhang2019robust, shenoi2020jrmot} firstly combine the camera with the Lidar sensor that is capable of offering precise spatial information. Leveraging the sensor fusion technology and redundant information from multiple sensors, the performance of 3D MOT can be significantly boosted. After that, a pairwise feature similarity between any two objects in the different frames is learned from fused features. Finally, the data association algorithms. For example, Hungarian algorithm \cite{kuhn1955hungarian}, Joint Probabilistic Data Association (JPDA) \cite{fortmann1983sonar} are employed to assign different IDs to corresponding objects for data association.

It is easy to observe that capturing the discriminative feature, which is beneficial for distinguishing different objects, is the critical  process when learning the affinity matrix and the following data association. One efficient method to solve this problem is to represent the objects in the adjacent frames as a directed graph. Specifically, each object can be treated as a node in the graph, and the relationship between an object pair is the edge between related two nodes. As a result, the problem of exploiting the discriminative feature between two objects is convert to learn the relations between the nodes in a directed graph. Consequently, we propose a deep neural network by employing the graph neural network into the 3D MOT to better exploit discriminative features.

Furthermore, considering the fact that the nodes in the graph are unordered, we are unlikely to leverage the advantages of the convolutional neural networks (CNNs) to exploit the features. Prior works for both 2D MOT and 3D MOT \cite{sun2019deep, wang2019towards, weng2020gnn3dmot, hu2019joint, zhang2019robust} use the multi-layer perceptron (MLP) to capture contextual features for each node. However, it is not efficient to learn local features between nodes using MLP, as the MLP operation is not a spatial convolution. As a result, we introduce a novel convolutional operation, named RelationConv, to better exploit relations between nodes in the graph.

In order to learn temporal-spatial features for the objects in the sequence, we also propose a feature extractor that jointly learns appearance features and motion features from both 2D images and 3D point clouds. Specifically, we use the off-the-shelf 2D image and 3D point cloud detectors to extract respective features for the appearance features of the objects. We learn the motion features by building a subnetwork that takes the 2D bounding boxes in the images and 2D bounding boxes in the point clouds as input. We finally fuse the appearance and motion features together for further data association. We summarize our contributions as:
\begin{itemize}
\item We represent the detected objects as the nodes in a directed graph and propose a graph neural network to exploit discriminative features for the objects in the adjacent frames for 3D MOT.
\item We propose a novel joint feature extractor to learn both 2D/3D appearance features and motion features from the images and the point clouds in the sequence.
\item We novelly propose a RelationConv operation to efficiently learn the correlation between each pair of objects for the affinity matrix.
\end{itemize}

\section{Related Work}
\subsection{2D Multi-Object Tracking.} 
The MOT methods can be categorized into the online method and offline method. The online method only predicts the data association between the detection in current frame and a few past frames. It is normally used for real-time applications. Early 2D trackers \cite{ma2015hierarchical, wang2015visual, bhat2018unveiling, danelljan2017eco, danelljan2016beyond} enhance the robustness and accuracy of tracking by exploiting and combining deep features in the different layers. However, these integrated features from multiple layers are not helpful when the targets are heavily occluded or even unseen in a certain frame. Several 2D MOT methods \cite{choi2017attentional, kiani2017learning, mueller2017context} employ correlation filters to improve the decision models. Deep reinforcement learning is used in \cite{yun2017action} to efficiently predict the location and the size for the targets in the new frame.

On the other hand, the offline method aims to find global optimization for the whole sequence. Some models \cite{zhang2008global, schulter2017deep} build a neural network with a min-cost flow algorithm to optimize the total cost for the data association problem.

\subsection{3D Multi-Object Tracking}
3D object detection has made a great success in recent years, especially when PointNets \cite{qi2017pointnet, qi2017pointnet++} are capable of processing the unordered point cloud in an efficient way. As a result, many researchers draw attention to 3D object tracking based on accurate 3D detection results. Some approaches \cite{weng2019baseline, osep2017combined, scheidegger2018mono} firstly predict the 3D objects from off-the-shelf 3D detectors, followed by a filter-based model to track the 3D objects continuously. \textit{mmMOT} \cite{zhang2019robust} builds an end-to-end neural network to extract features for the detected objects and data association. Specifically, it employs 2D feature extractor \cite{simonyan2014very} and 3D feature extractor \cite{qi2017pointnet} to capture 2D and 3D features for objects in the adjacent two frames. A sensor fusion module is then proposed to aggregate multi-modality features, which is finally used for further data association. However, the model only learns appearance features for the detected objects, and motion features are not considered. Alternatively, \textit{GNN3DMOT} \cite{weng2020gnn3dmot} proposes a joint feature extractor to learn discriminative appearance features for the objects from the images and the point clouds, and then employs an LSTM neural network to capture the motion information. Finally, a batch triplet loss is processed for data association.

\subsection{Joint Multi-Object Detection and Tracking}
Joint multi-object detection and tracking method becomes popular due to the fact that it might lead to a sub-optimal solution if the detection task and the tracking task are decoupled. Recent methods \cite{yin2020center, zhou2020tracking} construct an end-to-end framework with a multi-task loss to directly localize the objects and associate them with the objects in previous frame. Similarly, \textit{FaF} \cite{luo2018fast} firstly converts the sequential point clouds into stacked voxels, and then applies standard 3D CNN over 3D space and time to predict 3D objects location, associate them in the multiple frames, and forecast their motions.

\subsection{Data Association in MOT}
Data association is an essential problem in the MOT task to assign the same identity for the same objects in the sequence. Traditionally, the Hungarian algorithm \cite{kuhn1955hungarian} minimizes the total cost of the similarity for each pair of observations and hypotheses. \textit{JPDA} \cite{fortmann1983sonar} considers all the possible assignment hypotheses and uses a joint probabilistic score to associate objects in the different frames. Modern works \cite{butt2013multi, zhang2008global} firstly represent the objects and corresponding relations as a directed graph. Each object is treated as the node in the graph, and relation between each pair of objects is the edge. After that, the data association problem can be cast as a linear program problem for seeking the optimal solution.

\begin{figure*}[t!]
  \centering
   {\epsfig{file = 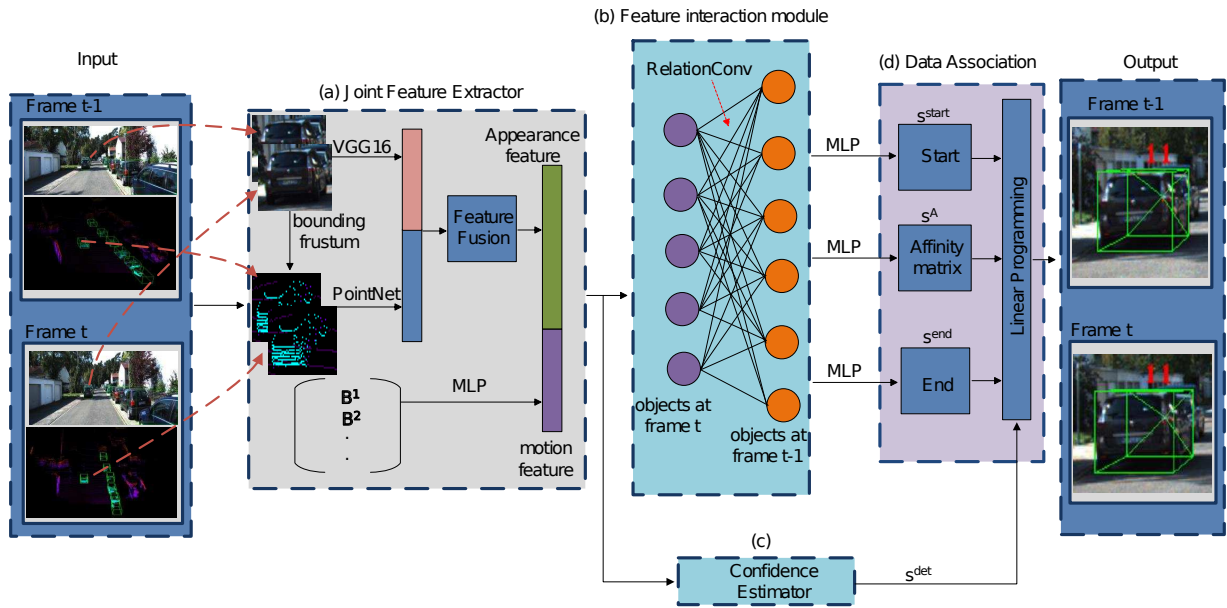, width=1\linewidth}}
  \caption{\textbf{Relation3DMOT architecture:} The architecture consists of 4 main parts. (a) takes the objects with 2D/3D bounding boxes as input, and then applies 2D feature extractor VGG-16 and 3D feature extractor PointNet on the cropped image and point cloud respectively to capture 2D/3D features, followed by a feature fusion module to obtain the appearance feature for each detected object. The motion feature is extracted by applying the aggregated 2D/3D bounding boxes on a proposed multi-layer perceptron (MLP) network. Finally, we concatenate the appearance feature and the motion feature to generate the fused global feature for each object. \(\left\{\mathbf{B_1, B_2, \ldots B_N}\right\}\) denotes a list of the aggregated 2D/3D bounding boxes information. (b) We propose a novel RelationConv operation to build a feature interaction module for discriminative feature extraction. (c) proposes a confidence estimator to predict whether the detected objects are true positive. (d) We finally learn an affinity matrix and predict several binary scores to optimize the data association problem.}
  \label{fig:structure}
 \end{figure*}

\section{Model Structure}
Our proposed 3D MOT network follows the tracking-by-detection paradigm. As shown in Figure~\ref{fig:structure}, the framework takes a sequence of images and related point clouds as input, and consists of three modules: (a) a joint feature extractor to capture the appearance feature and the motion feature from the 2D images and 3D point clouds; (b) a feature interaction module that takes the joint feature as input and uses proposed RelationConv operation to exploit the correlation between the pairs of objects in the different frames; (c) a confidence estimator predicts if a certain detected object is a valid detection; (d) a data association module to compute the affinity matrix for associating the objects in the adjacent frames.

\subsection{Problem Statement}
As an online 3D MOT method, our model performs objects association in every two consecutive frames from a given sequence. We refer to current frame at time \(t\) with \(N\) detected objects as \(\mathbf{X^t}=\left\{ \mathbf{{x_i}^{t}} |  i=1,2,\ldots,N\right\}\), and previous frame at time \(t-1\) with \(M\) detected objects as~\(\mathbf{X^{t-1}}=\left\{ \mathbf{{x_j}^{t-1}} |  j=1,2,\ldots,M\right\}\). We aims at exploiting discriminative feature for each \(\mathbf{{x_i}^{t}}\) and \(\mathbf{{x_j}^{t-1}}\) pair, predicting a feature affinity matrix for the correct matching, and finally assigning the matched IDs to corresponding objects in current frame \(t\).

\subsection{2D Detector}
For object detection, there are two methods to localize the objects in the frames. One method is to use existing 3D detectors (e.g. \textit{PointRCNN} \cite{shi2019pointrcnn}, \textit{PointPillar} \cite{lang2019pointpillars}) to predict 3D bounding boxes, which are then projected to corresponding images to obtain 2D bounding boxes. However, we believe the fact that the point clouds are lack of rich colour and texture information that allows to provide sufficient semantic information for the objects. As a result, it is still not efficient to localize the objects precisely. The other method is to use off-the-shelf 2D detectors to predict 2D bounding boxes. With respect to the 3D location, we employ \cite{qi2018frustum} to estimate the localization of the objects. We use this method and introduce \textit{RRC-Net} \cite{ren2017accurate} as our 2D detector model, as we observe that it could obtain higher recall and accuracy.

\subsection{Joint Feature Extractor}
In order to exploit sufficient information for the detected objects in the frames, we propose a joint feature extractor (see Figure~\ref{fig:structure} (a)) to learn deep representations from both the images and the point clouds. Specifically, we employ a modified \textit{VGG-16} \cite{simonyan2014very} to extract 2D appearance features, and then apply \textit{PointNet} \cite{qi2017pointnet} over the trimmed points, which are obtained by extruding related 2D bounding boxes into the 3D frustums in the point cloud, to capture 3D appearance features and predict 3D bounding boxes. After that, we propose a sensor fusion module to aggregate the 2D and 3D appearance features together for further feature interaction. With regard to the motion features, we build a subnetwork to learn the high-level motions of objects using the information of the 2D bounding boxes and 3D bounding boxes.

\subsubsection{2D Appearance Feature Extraction} 
\begin{figure*}[t!]
  \centering
   {\epsfig{file = 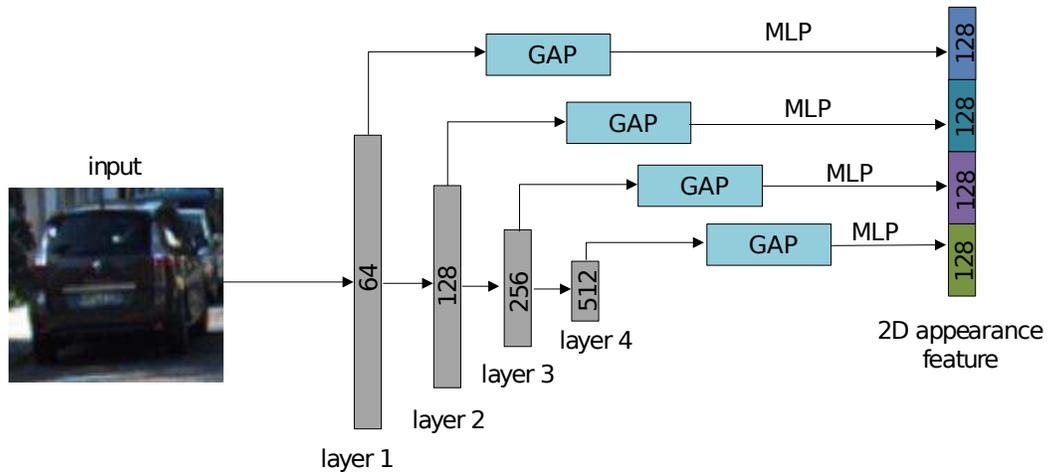, width=0.85\linewidth}}
  \caption{\textbf{Modified VGG-16:} The module takes the cropped and resized object patches with dimension \(224 \times 224\) as input, and the layers output 64, 128, 256, 512 channels of the features respectively after max pooling operation. After that the global average pooling (GAP) operation and multi-layer perceptron (MLP) are used to obtain semantic features with 128 channels for each layer. Finally, all the layers are concatenated together to generate the global feature that contains the information in all the layers.}
  \label{fig:vgg}
 \end{figure*}

As shown in Figure~\ref{fig:structure} (a), we take the objects in the 2D bounding boxes in the image as input, which are then cropped and resized to the fixed size \(224 \times 224\) to adapt the \textit{VGG-16} \cite{simonyan2014very} feature extractor. \cite{wang2015visual} indicates that the features in the different CNN layers contain different semantic properties. For example, a lower convolutional layer is likely to capture more detailed spatial information but low-level features, and a deeper convolutional layer could capture more abstract and high-level information. As a result, inspired by \cite{zhang2019robust, wang2015visual}, we embed a skip-pool \cite{bell2016inside} method into \textit{VGG-16} network (see Figure~\ref{fig:vgg}) to involve all the features in the different levels for the global feature generation, which is treated as the 2D appearance feature for the corresponding detected object.

\subsubsection{3D Appearance Feature Extraction} 
We firstly obtain the 3D points of each object by extruding the corresponding 2D bounding box into a 3D bounding frustum, where the 3D points on the object are then trimmed from. After that, we apply the \textit{PointNet} \cite{qi2017pointnet} to capture the spatial feature for 3D appearance feature and predict corresponding 3D bounding box.

\subsubsection{Motion Feature}
With respect to the motion feature for each object, we directly use the 2D bounding box and corresponding 3D bounding box as the motion cue. We define the 2D bounding box of a certain object as \(\mathbf{B_{2d}}=[x_{2d},y_{2d},w_{2d},h_{2d}]\), where \([x_{2d}, y_{2d}]\) is the center of the 2D bounding box, and \(w_{2d}, h_{2d}\) are the width and the height of the 2D bounding box respectively. Similarly, the 3D bounding box is defined as \(\mathbf{B_{3d}}=[x_{3d},y_{3d},z_{3d},w_{2d},h_{2d},l_{2d},{\theta}_{2d}]\), where \([x_{3d}, y_{3d}, z_{3d}]\) is the center of the 3D bounding box, and \(w_{3d}, h_{3d}, l_{3d}\) are the width, height, length of the 3D bounding box respectively. \({\theta}_{3d}\) indicates the orientation of the object in the 3D space. Finally, the motion cue is defined as \(\mathbf{B=[B_{2d}, B_{3d}]}\) (see Figure~\ref{fig:structure} (a)), which is then fed to a 3-layer MLP subnetwork to capture the motion feature for each detected object.

\subsubsection{Features Aggregation and Fusion} 
We firstly aggregate 2D appearance feature and 3D appearance feature to exploit sufficient semantic features (e.g. spatial and colour features) for each detected object. We propose three fusion operators: (1) An intuitive method is to add 2D appearance feature and 3D appearance feature together after forcing them to have the same feature channels; (2) Another common approach is to concatenate 2D appearance feature and 3D appearance feature, after which a 1-layer MLP is used to adapt the dimension of the fused features; (3) Another operator is attention-based weighted sum. Specifically, a scalar score is firstly computed by learning a 1-dimension feature from the features in the different sensors, after which the score is normalized by computing a sigmoid function. Finally, the weighted sum for the fusion is computed by using an element-wise multiplication operation on the normalized scores and corresponding features obtained from different sensors. We compare the performance by applying different fusion operators on our model in the experiments, and using a concatenation operator as our 2D and 3D appearance features aggregation method in our model.

We finally fuse the appearance feature and motion feature by concatenating the aggregated 2D/3D appearance feature with the related motion feature for further feature interaction.

\subsection{Feature Interaction Module}
After the joint feature extractor, we gain the fused features for \(M\) individual objects in the frame \(t-1\) and \(N\) objects in the frame \(t\). We then propose a feature interaction module, as shown in Figure~\ref{fig:structure} (b), to learn the relations between each pair of two objects (one object is in previous frame \(t-1\) and the other object is in current frame \(t\)).

\subsubsection{Graph Construction}
In order to efficiently represent the objects in the different frames, we treat the feature of each object as a node in a graph structure, and the edge between two nodes in the graph can indicate the relationship between two objects in the different frames. As a result, we firstly construct a directed acyclic graph structure to represent the objects in two adjacent frames. Generally, a directed acyclic graph can be defined as~\(\mathbf{G=(V,E)}\), where \(\mathbf{V} \subset \mathbb{R}^F\) are the nodes with \(F\) dimension, and \(\mathbf{E \subseteq V \times V}\) are edges connecting two nodes in the graph.

However, considering the fact that we only learn the correlation between every object in the frame \(t\) and all the objects in the frame \(t-1\), rather than taking the relations between the pair of objects in the same frame into account, our graph for the objects representation is constructed as~\(\mathbf{G=(V^t,E^t)}\), \(\mathbf{E^t \subseteq V^t \times V^{t-1}}\), where \(\mathbf{V^t}\) indicate \(N\) objects at current frame \(t\), and \(\mathbf{V^{t-1}}\), which are also the neighbourhood nodes of \(\mathbf{V^t}\), denote \(M\) objects at previous frame \(t-1\). The edge feature set \(\mathbf{E^t}\) for all the nodes in current frame are defined as Equations~\ref{eq:edge}:

\begin{equation}\label{eq:edge}
\begin{split}
    &\mathbf{X^t}=\left\{ \mathbf{{x_i}^t} \in \mathbb{R}^F, i=1,2,\ldots,N\right\}\\
    &\mathbf{V^t = X^t}\\
    &\mathbf{{{y_i}_j}^t=|{x_i}^t-{{x_j}^{t-1}}|}, j=1,2,\ldots,M\\
    &\mathbf{{e_i}^t=({{y_i}_1}^t,{{y_i}_2}^t,\ldots,{{y_i}_M}^t)} , i=1,2,\ldots,N\\
    &\mathbf{E^t} = \left\{ \mathbf{{e_i}^t} | i=1,2,\ldots,N\right\}
\end{split}
\end{equation}
where \(\mathbf{{x_i}^t}\) indicates a certain object node in current frame \(t\), \(\mathbf{{{y_i}_j}^t}\) denotes the edge feature connecting \(i\)-th object node \(\mathbf{{x_i}^t}\) in current frame \(t\) and \(j\)-th object node \(\mathbf{{x_j}^{t-1}}\) in previous frame \(t-1\). The \(|\cdot|\) is the absolute value operation. \(\mathbf{{e_i}^t}\) are the edge features connecting \(i\)-th object node in current frame \(t\) and all the \(M\) neighbourhood nodes in previous frame \(t-1\).

\subsubsection{Relation Convolution Operator}
Considering irregular and unordered properties of the nodes in the graph, we are unlikely to use regular CNN filters (e.g. \(3 \times 3,  7 \times 7 filters\)) on the unstructured graph for convolutional operation, as these filters are only suited to deal with the standard grid data, such as images. As a result, the traditional method is to apply shared MLP on the graph to learn local and global contextual representations. However, it is not efficient to extract the spatially local features from unordered data using the shared \(1 \times 1\) filters, which leads to small receptive fields. Inspired by the processing of standard convolutional kernels, \textit{PointCNN} \cite{li2018pointcnn} learned a transformation matrix as a regular CNN filter to capture local features for the irregular point clouds, and it outperforms MLP-based methods \textit{PointNets} \cite{qi2017pointnet, qi2017pointnet++} by a large margin.

In order to leverage the ability of CNNs that is capable of extracting spatially-local correlation with large receptive fields, we propose a relation convolution operator, named RelationConv, to abstract fine-grained local representations for the nodes on the graph. The advantages of our RelationConv are that it is similar to standard convolutional kernels and can work on irregular data (e.g. graphs). In the standard CNNs, the convolutional operation between the filters \(\mathbf{W}\) and the feature map \(\mathbf{X}\) can be defined as Equation \ref{eq:conv} to obtain abstract features \(\mathbf{f_X}\).

\begin{equation}\label{eq:conv}
    \mathbf{f_X = W \cdot X}
\end{equation}
where \(\mathbf{X}\) represents the feature map with standard grid distribution, and \(\mathbf{W}\) are the convolutional filters. The operation "\(\mathbf{\cdot}\)" denotes element-wise multiplication.

\begin{figure*}[!t]
  \centering
   {\epsfig{file = 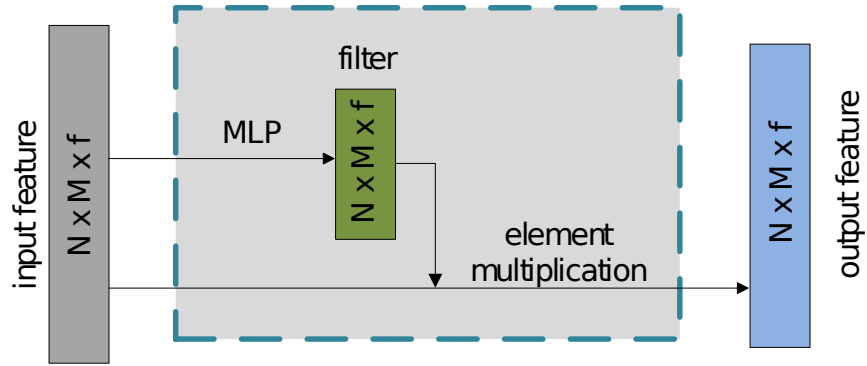, width=0.7\linewidth}}
  \caption{RelationConv Operation.}
  \label{fig:relationconv}
 \end{figure*}

Similarly, as defined in Equation \ref{eq:relationconv}, our RelationConv firstly learns a flexible filter \(\mathbf{W()}\) by applying a shared MLP with a non-linear function on the edge feature of the unordered graph, and then an element-wise multiplication operation is used on the learned filter \(\mathbf{W}\) and the edge feature \(\mathbf{E^t}\) to extract the local feature for the nodes in the graph (see Figure \ref{fig:relationconv}). It is easy to observe that our flexible filter \(\mathbf{W}\) is learned from all the nodes and corresponding neighbourhood in the graph, which forces it to consider the global information in the graph, and also makes it independent to the ordering of the nodes.

\begin{equation}\label{eq:relationconv}
\begin{split}
    &\mathbf{W(E^t) = RELU(MLP(E^t))}\\
    &\mathbf{f_{E^t} = W(E^t) \cdot E^t}\\
\end{split}
\end{equation}
where \(\mathbf{E^t}\) is the edge feature set. \(\mathbf{W(E^t)}\) are the learnable and flexible filters obtained from a shared Multi-layer perceptron \(MLP()\) with a non-linear function \(RELU()\) \cite{xu2015empirical}. \(\mathbf{f_{E^t}}\) are the abstract features captured from our RelationConv operation.

\subsubsection{Feature Interaction}
We believe that the same objects in the different frames should learn similar discriminative features, and the feature similarity should be dependent. In contrast, the feature similarity of two different objects should be decreased. As a result, the discriminative feature is beneficial for avoiding confusion while matching the objects.

Given the obtained fused features containing 2D/3D appearance features and motion features, we propose a feature interaction module to learn the correlation between each object pair in the different frames as shown in Figure \ref{fig:structure} (b). Rather than directly learning the deep affinity for further data association, we firstly employ a feature interaction module equipped with a 1-layer RelationConv network to learn more discriminative features, which allows the feature communication between two objects in adjacent frames before learning the affinity matrix. 

\subsubsection{Confidence Estimator}
As shown in Figure \ref{fig:structure} (c), we build a 3-layer MLP classification network to predict the scalar scores \(\mathbf{s^{det}}\) as the confidence of the validation for the detected objects.

\subsection{Data Association}
\subsubsection{Affinity Matrix Learning}
In order to associate and match the objects in the different frames, given the correlation representation after feature communication among objects, we use a 3-layer MLP subnetwork to learn an affinity matrix with 1-dimension output feature \(\mathbf{s^A}\). The affinity matrix is capable of determining whether a certain object pair indicates a link. Besides, the scalar score in the matrix shows the confidence of the object pair associating with the same identity.

Furthermore, inspired by \cite{zhang2019robust}, we further learn a start estimator and an end estimator to predict whether an object is linked. Specifically, the start estimator learns the scalar scores \(\mathbf{s^{start}}\) to determine whether a certain object just appears in previous frame \(t-1\). On the other hand, the end estimator predicts the score \(\mathbf{s^{end}}\) whether a certain object is likely to disappear in the frame \(t\) due to hard occlusion or out-of-bounds, etc. The start estimator and the end estimator firstly use an average pooling over the deep correlation representation to summarize the relations, and then employ the respective MLP network to learn the scalar scores for all the objects.

\subsubsection{Linear Programming}
We obtain several binary variables for prediction scores from our proposed neural network as shown in Figure~\ref{fig:structure} (d),. In summary, the detection score \({s_i}^{det}\) indicates the confidence whether \(i\)-th object is a true positive detection. \({{s_i}_j}^A\) denotes the affinity confidence whether the \(j\)-th object in previous frame \(t-1\) and the \(i\)-th object in current frame \(t\) are the same objects. \({s_j}^{start}\) denotes the confidence whether \(j\)-th object in previous frame \(t-1\) starts a new trajectory in frame \(t-1\). \({s_i}^{end}\) denotes the confidence whether \(i\)-th object in current frame \(t\) ends a trajectory in the frame \(t\). We then aggregate all the prediction scores to a new vector \(\mathbf{S}=[{s_i}^{det}, {{s_i}_j}^A, {s_j}^{start}, {s_i}^{end}]\) for the optimization of the data association problem.

Considering the graph structure for all the detected objects, the data association problem can be formulated as the min-cost flow graph problem \cite{zhang2019robust, schulter2017deep}. Specifically, we use these obtained prediction scores to define linear constraints, and then find an optimal solution for matching problem.

There are two circumstances for a certain true positive object in previous frame \(t-1\). It can be matched to another object in current frame \(t\), or it starts a new trajectory. As a result, we define the linear constraint as Equation \ref{eq:constraint1}.

\begin{equation}\label{eq:constraint1}
    {s_j}^{det} = {s_j}^{start} + \sum_{i=0}^{N}{{{s_i}_j}^A}
\end{equation}

Similarly, a certain true positive object in current frame \(t\) can be matched to another object in previous frame \(t-1\), or it ends a trajectory. Consequently, the linear constraint can be defined as Equation \ref{eq:constraint2}.

\begin{equation}\label{eq:constraint2}
    {s_i}^{det} = {s_i}^{end} + \sum_{j=0}^{M}{{{s_i}_j}^A}
\end{equation}

Finally, we formulate the data association problem as:

\begin{equation}\label{eq:constraint3}
\begin{split}
    &\mathop{\arg\min}_{s} = \mathbf{\Theta(X)^\mathrm{T}S}\\
    &s.t.\quad  \mathbf{CS} = 0, \mathbf{S} \in \left\{0, 1\right\}\\
\end{split}
\end{equation}
where \(\mathbf{\Theta(X)}\) indicates a flattened vector that comprises all the prediction scores. \(\mathbf{CS}\) is a matrix form that satisfies two linear constraint Equations \ref{eq:constraint1} and \ref{eq:constraint2}.

\section{Experiments}
\subsection{Dataset}
We evaluate our neural network on the KITTI object tracking benchmark \cite{geiger2013vision, geigerwe}. The benchmark consists of 21 training sequences and 29 testing sequences. We split the training sequences into 10 sequences for training and 11 sequences for validation. As a result, we obtain 3975 frames for training and 3945 frames for validation. 
The dataset is captured from a car equipped with two color/gray stereo cameras, one Velodyne HDL-64E rotating 3D laser scanner and one GPS/IMU navigation system. Each object is annotated with a unique identity number (ID) across the frames in the sequences, 2D bounding boxes 3D bounding boxes parameters. We measure the distance between the predicted bounding box and corresponding bounding box of matched object-hypothesis by calculating the intersection over union (IoU).

\subsection{Evaluation Metrics}
The evaluation metrics to assess the performance of tracking methods are based on CLEAR MOT \cite{bernardin2008evaluating} and \cite{li2009learning}. Specifically, MOT precision (MOTP) measures the average total error of distances for all the frames in the sequences as defined in Equation \ref{eq:motp}, and it indicates the total misalignment between the predicted bounding boxes and corresponding matched object-hypothesis.

\begin{equation}\label{eq:motp}
    MOTP = \frac{\sum_{i,t} {{d_t}^i}}{\sum_{t} {c_t}}
\end{equation}
where \({d_t}^i\) indicates the distance between the bounding box of \(i\)-th object and corresponding matched hypothesis in the frame \(t\). \(c_t\) indicates the total number of the matched objects in the frame \(t\).

MOT accuracy (MOTA) measures the total tracking accuracy for all the frames as defined in Equation \ref{eq:mota}.

\begin{equation}\label{eq:mota}
    MOTA = 1-\frac{\sum_{t} {({FN}_t+{FP}_t+{IDSW}_t})}{\sum_{t} {GT}_t}
\end{equation}
where \({FN}_t, {FP}_t, {IDSW}_t, {GT}_t\) denote the total number of false negative objects, false positive objects, identity switches and ground truth respectively in the frame \(t\).

Besides, \cite{li2009learning} introduces other metrics to improve the tracking assessment, such as mostly tracking (MT) indicating the percentage of the entire trajectories in the sequences that could cover more than 80\% in total length; mostly lost (ML) indicating the percentage of the entire trajectories in the sequences that could cover less than 20\% in total length; partial tracked (PT) indicating \(1-MT-ML\).

\subsection{Training Settings}
Our model is implemented using Pytorch 1.1. We train our model on a ThinkStation P920 workstation with one NVIDIA GTX 1080Ti, and use Adam as a training optimization strategy with the initial learning rate 3e-4. Besides, the super convergence training strategy is employed to boost the training processing and the maximum learning rate is set to 6e-4.

\subsection{Results}
Table~\ref{tab:trk} shows that our Relation3D model achieves competitive results compared to recent state-of-the-art online tracking methods. It is easy to observe that our model achieves the best ID-SW metric when we compare to 2D online tracking methods. We discuss that 3D spatial information and location are capable of avoiding the confusion when matching the pairwise objects. Our MOTA and MOTP results outperform those of \textit{GNN3DMOT} \cite{weng2020gnn3dmot} whose feature interaction mechanism employs MLP network to exploit discriminative features. It shows the effectiveness of our RelatiionConv operation for feature interaction. Compared to \textit{mmMOT} \cite{zhang2019robust}, our model is slightly better as it is beneficial from our RelationConv operation and the combination of both the appearance features and motion features.

\begin{table}[t!]
  \caption{3D tracking results on KITTI tracking test set.}
  \label{tab:trk} \centering
  \begin{tabular}{cccccccc}
    \toprule[0.96pt]
    Methods  &Input &  MOTA(\%) & MOTP(\%) & ID-SW  &Frag  & MT  & ML\\
    \midrule
    BeyondPixels \cite{sharma2018beyond}     &2D               & 84.24 & \textbf{85.73} & 468     & 944  & 73.23 & \textbf{2.77} \\
    3DT \cite{hu2019joint}     &2D               & 84.52 & 85.64 & 377     & 847  & 73.38 & \textbf{2.77} \\
    MASS \cite{karunasekera2019multiple}     &2D               & \textbf{85.04} & 85.53 & 301     & 744  & \textbf{74.31} & \textbf{2.77} \\
    \midrule
    AB3DMOT \cite{weng20203d}     &3D               & 83.84 & 85.24 & \textbf{9}     & \textbf{224}  & 66.92 & 11.38 \\
    \midrule
    FANTrack \cite{baser2019fantrack}      &2D + 3D             & 77.72 & 82.32   & 150   & 812  & 62.61 & 8.76  \\
    mmMOT \cite{zhang2019robust}        &2D + 3D              & 84.43 & 85.21  & 400   & 859  & 73.23 & \textbf{2.77}  \\
    GNN3DMOT \cite{weng2020gnn3dmot}  &2D + 3D                 & 80.40 & 85.05   & 113   & 265  & 70.77 & 11.08 \\
    \midrule
    OURS           &2D + 3D            & 84.78 & 85.21  & 281   & 757  & 73.23 & \textbf{2.77} \\
    \bottomrule[0.96pt]
  \end{tabular}
\end{table}

\begin{figure*}[t!]
  \centering
   \subfigure{\includegraphics[width=0.7\linewidth]{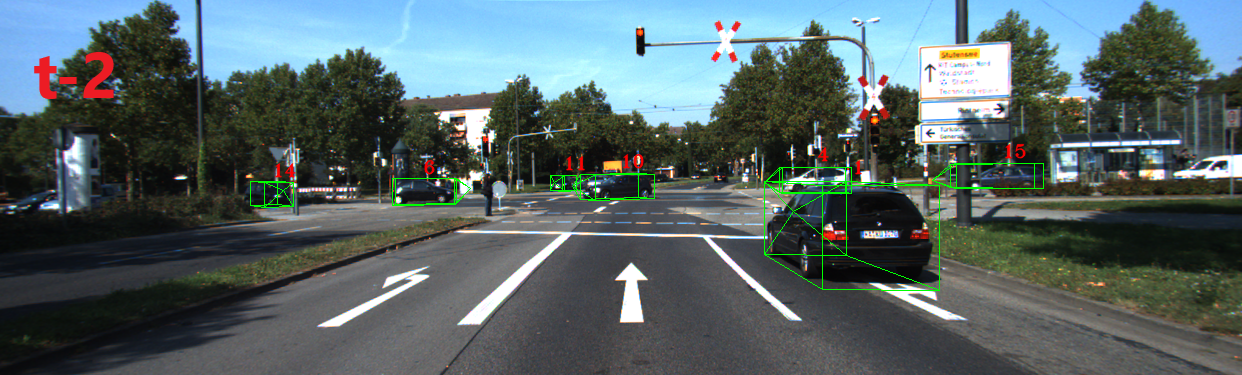}} 
   \subfigure{\includegraphics[width=0.7\linewidth]{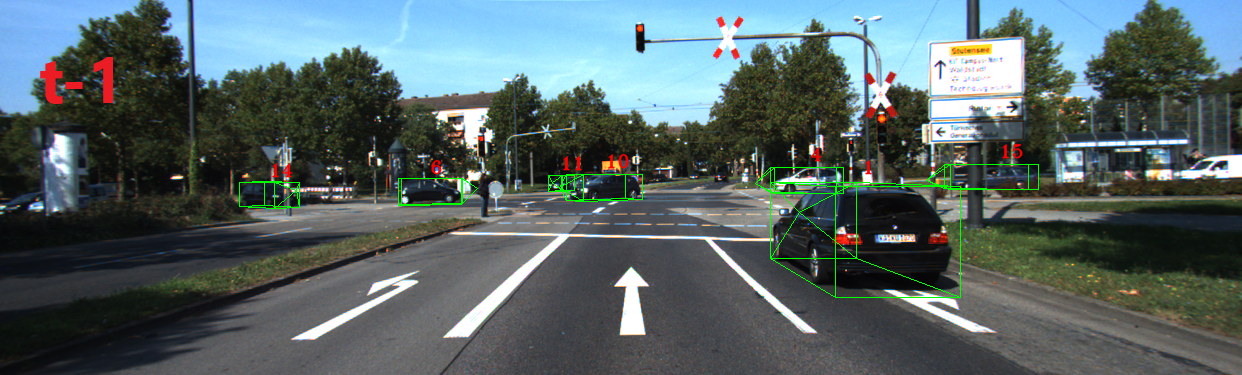}} 
   \subfigure{\includegraphics[width=0.7\linewidth]{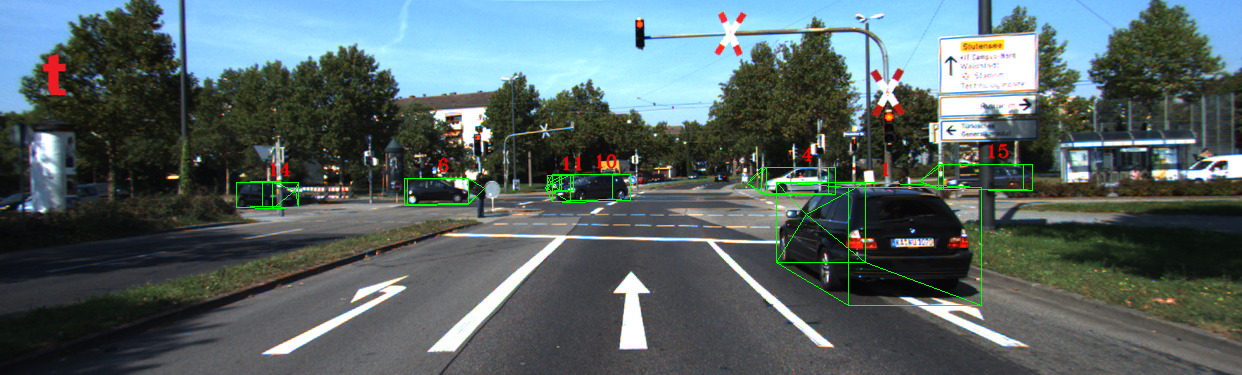}} 
  \caption{Qualitative results on the KITTI object tracking benchmark validation dataset.}
  \label{fig:visu}
 \end{figure*}

\subsection{Ablation study}
We investigate different hyper-parameter settings to evaluate the effectiveness of our model on the KITTI object tracking benchmark \cite{geiger2013vision}.

Table \ref{tab:fusion} indicates the effectiveness of different fusion methods. It shows that the concatenation operation outperforms addition and attention-based weighted sum methods. We believe that the concatenation operation is capable of exploiting more useful information from the features obtained from different sensors. Specifically, the addition operation is inefficient to align the features captured from 2D RGB images and 3D point clouds, making it difficult to learn discriminative feature after element-wise addition. Although the attention-based weighted sum method enables to highlight the importance for the different features, the concatenation operation is a more general operation that gathers all the information from different modalities.

Table \ref{tab:edge} shows that the performance is the best when the edge feature \({{y_i}_j}^t=|{x_i}^t-{{x_j}^{t-1}}|\). We observe that the first option \({x_j}^{t-1}\) only considers the neighbourhood information without involving then center nodes. The third option \({x_i}^t - {x_j}^{t-1}\) uses the relative distance value between the object pair as the edge feature, which is efficient when we deal with spatial data (e.g. point cloud). However, the result represents that the absolute value is more efficient when we learn discriminative features for the similarity. The fourth option \([{x_i}^t, {x_j}^{t-1}]\) encodes the edge feature by combining the individual information without explicitly considering the distance between each pair of objects.

We also investigate the effectiveness of the appearance feature and motion feature as shown in Table \ref{tab:feature}, which indicates that the motion feature could contribute the \(MOTA\) by 1.07\%, and significantly reduce the number of \(ID-SW\) and \(Frag\) by 119 and 113 respectively, which convincingly verifies that the motion of 2D/3D localization could help to match the correct objects for data association.

Table \ref{tab:conv} compares our RelationConv operation with traditional MLP method for local feature extraction. It shows that our RelationConv outperforms MLP method by significant 1.2\% for MOTA metric and performs much better in terms of \(ID-SW\) and \(Frag\) metrics.

It is worthwhile noting that the \(MOTP\) metric is only related to the distance between predicted bounding boxes and corresponding matched object-hypothesis as shown in Equation \ref{eq:motp}. Besides, the \(MT\) and the \( ML\) are irrelevant to the measurement of whether the IDs of the objects remain the same throughout the entire sequence. As a result, the metrics of \(MOTP, MT, ML\) are unlikely to update during our ablation investigation (see Table \ref{tab:fusion}, \ref{tab:edge}, \ref{tab:feature}, \ref{tab:conv} as long as the performance of object detection remains the same.

\begin{table}[t!]
  \caption{Effectiveness of different fusion methods.}
  \label{tab:fusion} \centering
  \begin{tabular}{ccccccc}
    \toprule[0.96pt]
    Fusion method   &  MOTA (\%) & MOTP (\%) & ID-SW  &Frag  & MT (\%)  & ML (\%)\\
    \midrule
    add               & 91.72 & 90.35 & 106     & 210  & 90.28 & 0.9 \\
    concatenate               & \textbf{92.33} & \textbf{90.35}  & \textbf{38}   & \textbf{143}  & \textbf{90.28} & \textbf{0.9} \\
    weighted sum               & 91.94 & 90.35 & 82     & 187  & 90.28 & 0.9 \\
    \bottomrule[0.96pt]
  \end{tabular}
\end{table}

\begin{table}[t!]
  \caption{Effectiveness of different edge feature aggregation operations.}
  \label{tab:edge} \centering
  \begin{tabular}{ccccccc}
    \toprule[0.96pt]
    Edge Feature   &  MOTA (\%) & MOTP (\%) & ID-SW  &Frag  & MT (\%)  & ML (\%)\\
    \midrule
    \({x_j}^{t-1}\)               & 90.94 & 90.35 & 193     & 290  & 90.28 & 0.9 \\
    \(|{x_i}^t - {x_j}^{t-1}|\)               & \textbf{92.33} & \textbf{90.35}  & \textbf{38}   & \textbf{143}  & \textbf{90.28} & \textbf{0.9} \\
    \({x_i}^t - {x_j}^{t-1}\)               & 91.85 & 90.35 & 92     & 194  & 90.28 & 0.9 \\
    \([{x_i}^t, {x_j}^{t-1}]\)               & 91.96 & 90.35 & 80     & 181  & 90.28 & 0.9 \\
    \bottomrule[0.96pt]
  \end{tabular}
\end{table}

\begin{table}[t!]
  \caption{Effectiveness of appearance feature (A), motion feature (M) and RelationCov operation.}
  \label{tab:feature} \centering
  \begin{tabular}{ccccccc}
    \toprule[0.96pt]
    Feature   &  MOTA (\%) & MOTP (\%) & ID-SW  &Frag  & MT (\%)  & ML (\%)\\
    \midrule
    A               & 91.26 & 90.35 & 157     & 256  & 90.28 & 0.9 \\
    A + M            & \textbf{92.33} & \textbf{90.35}  & \textbf{38}   & \textbf{143}  & \textbf{90.28} & \textbf{0.9} \\
    \bottomrule[0.96pt]
  \end{tabular}
\end{table}

\begin{table}[t!]
  \caption{Effectiveness of different convolutional operations.}
  \label{tab:conv} \centering
  \begin{tabular}{ccccccc}
    \toprule[0.96pt]
    Method   &  MOTA (\%) & MOTP (\%) & ID-SW  &Frag  & MT (\%)  & ML (\%)\\
    \midrule
    MLP               & 91.13 & 90.35 & 171     & 271  & 90.28 & 0.9 \\
    RelationConv               & \textbf{92.33} & \textbf{90.35}  & \textbf{38}   & \textbf{143}  & \textbf{90.28} & \textbf{0.9} \\
    \bottomrule[0.96pt]
  \end{tabular}
\end{table}

\section{Conclusion}
We propose a deep affinity network, named Relation3DMOT, to learn discriminative features and associate the objects in the adjacent frames for 3D MOT. We employ a joint feature extractor to capture the 2D/3D appearance feature and motion feature from 2D images and 3D point clouds respectively, followed by a feature interaction module to enhance the feature communication among objects in the different frames. We also propose an efficient convolutional operation, named RelationConv, to abstract semantic and contextual relations for each object pair. We finally perform extensive experiments on the KITTI object tracking benchmark to demonstrate the effectiveness of our Relation3DMOT tracker.

In the further, we plan to improve our model by considering the objects with hard occlusion in the frames. Furthermore, it would be worthwhile to develop an end-to-end framework for the joint task of object detection and tracking, and we believe that it could avoid the decoupling issue when we deal with the object detection and tracking separately.

\authorcontributions{Conceptualization, C.C.; methodology, C.C.; software, C.C.; validation, C.C., L.Z.F.; formal analysis, C.C.; investigation, C.C. and L.Z.F.; resources, A.T.; data curation, C.C.; writing--original draft preparation, C.C.; writing--review and editing, C.C., L.Z.F. and A.T.; visualization, C.C.; supervision, L.Z.F., A.T.; project administration, L.Z.F.; funding acquisition, A.T. All authors have read and agreed to the published version of the manuscript.}


\conflictsofinterest{The authors declare no conflict of interest.}

\reftitle{References}



\externalbibliography{yes}





\end{document}